\documentclass[11pt]{article}
\usepackage{geometry}
\usepackage{coling2020}
\usepackage{times}
\usepackage{url}

\usepackage{hyperref}
\usepackage{makecell}
\usepackage{latexsym}
\usepackage{microtype}
\usepackage{floatrow}
\usepackage{textcomp}
\usepackage{enumitem}
\usepackage{graphicx}
\usepackage{multirow}

\newfloatcommand{capbtabbox}{table}[][\FBwidth]

\colingfinalcopy

\title{SemEval-2020 Task 4: Commonsense Validation and Explanation}

\author{Cunxiang Wang\textsuperscript{1,2\thanks{Equal contribution}}, Shuailong Liang\textsuperscript{3\footnotemark[1]}, Yili Jin\textsuperscript{4}, Yilong Wang\textsuperscript{1}, \\ \textbf{Xiaodan Zhu\textsuperscript{5} and Yue Zhang\textsuperscript{2}}\\
\textsuperscript{1}Zhejiang University, China;
\textsuperscript{2}School of Engineering, Westlake University, China\\
\textsuperscript{3}Singapore University of Technology and Design, Singapore;\\
\textsuperscript{4}Sun Yat-Sen University, China;
\textsuperscript{5}Queen's University, Canada\\
  {\tt {wangcunxiang, zhangyue}@westlake.edu.cn}, \\
  {\tt shuailong\_liang@mymail.sutd.edu.sg}, 
  {\tt zhu2048@gmail.com}\\
  }

\date{}

\begin{document}
\maketitle
\begin{abstract}
  In this paper, we present SemEval-2020 Task 4,~\textbf{Com}monsense \textbf{V}alidation and \textbf{E}xplanation (\textbf{ComVE}), which includes three subtasks, aiming to evaluate whether a system can distinguish a natural language statement that \textit{makes sense} to humans from one that does not, and provide the reasons.
  Specifically, in our first subtask, the participating systems are required to choose from two natural language statements of similar wording the one that \textit{makes sense} and the one does not. The second subtask additionally asks a system to select the key reason from three options why a given statement does not make sense. In the third subtask, a participating system needs to generate the reason. We finally attracted 39 teams participating at least one of the three subtasks. For Subtask A and Subtask B, the performances of top-ranked systems are close to that of humans. However, for Subtask C, there is still a relatively large gap between systems and human performance. 
  The dataset used in our task can be found at \textit{https://github.com/wangcunxiang/SemEval2020-Task4-Commonsense-Validation-and-Explanation}; The leaderboard can be found at \textit{https://competitions.codalab.org/competitions/21080\#results}.
\end{abstract}

%The leaderboard is available at: %\textit{https://competitions.codalab.org/competitions/21080\#results}.

\section{Introduction}
\blfootnote{ This work is licensed under a Creative Commons Attribution 4.0 International License. License details: http://creativecommons.org/licenses/by/4.0/.}
%Natural Language Understanding (NLU) has received increasing research attention in recent years. % \cite{deng2018deep}.
%With language models trained on large corpora \cite{ELMO,BERT}, algorithms show better performance than humans on some benchmarks \cite{BERT,RoBERTa}.
In the past decades, computer' ability in processing natural language has significantly improved. However, its intelligence for understanding common sense expressed in language is still limited.
For example, it is straightforward for humans to judge that the following sentence is plausible, or makes sense: ``\emph{John put a turkey into a fridge}" while ``\emph{John put an elephant into the fridge}" does not, but it is non-trivial for a computer to tell the difference. Arguably, commonsense reasoning plays a central role in a natural language understanding system \cite{logical}. It is essential to gauge how well computers can understand whether a given statement \textit{makes sense}. In our task, we take an operational definition of~\textit{making sense} by asking human subjects to generate natural language statements that obey or violate their commonsense knowledge about the world. \footnote{Note that the term ``makes sense" may also be used to describe if a statement is meaningful, regardless of whether it is plausible or conforms to common sense. In this task the term refers to statements that obey common sense.}

Many existing tasks embed the evaluation of commonsense understanding in other tasks such as co-reference resolution \cite{WSC2012,WSC2015}, subsequent event prediction \cite{COPA}, ordinal common-sense inference  \cite{JOCI}, situations with adversarial generations \cite{SWAG}, event validation \cite{wang2018modeling}, reading comprehension \cite{RocStories2016,SemEval-2018-Task-11,MCScript}, dialogue \cite{mutual} and QA \cite{SQUABU,COMMONSENSEQA,OpenBookQA}. They verify whether a system is equipped with common sense by testing whether the system can give a correct answer when the input does not contain such knowledge. The above tasks do not directly evaluate commonsense validation and they do not explicitly identify the key factor required in a commonsense validation process. 

The SemEval-2020 Task 4 includes three subtasks on testing whether a system can distinguish natural language statements that make sense from those that do not, and probe the reasons.
In the first subtask, a system needs to choose the against-common-sense statement from two natural language statements of similar wordings, e.g., ``\textit{John put an elephant into the fridge}" and ``\textit{John put a turkey into the fridge}", respectively. The second task aims to find the key reason from three provided options why a given nonsensical statement does not make sense. For example, for the nonsensical statement, ``\textit{John put an elephant into the fridge}", the three options are ``\textit{An elephant is much bigger than a fridge}", ``\textit{Elephants are usually white while fridges are usually white}", and ``\textit{An elephant cannot eat a fridge.}" A system needs to identify the correct reason. In addition, the third task requires the participating systems to generate the reason automatically. We hope that the task and datasets can facilitate studies on commonsense validation, its interpretability, and the related natural language understanding and generation problems.%希望大家用我们的，具体的impact， plausibility 和 generation 结合 可解释性、生成的作用

There are 39 teams submitting valid systems to at least one subtask. In Subtask  A and Subtask B, top-performing systems achieve performances closed to that of human subjects. However, for Subtask C, there is still a relatively large between system and human performances.

%To our knowledge, our dataset has the most direct decision-making process in commonsense reasoning and is the first one asking reasons behind the deci\usepackage{}sion making process.\usepackage{}
\section{Task Setup}

\subsection{Task Definition}
Formally, each instance in our dataset is composed of eight sentences: \{$s_1$, $s_2$, $o_1$, $o_2$, $o_3$, $r_1$, $r_2$, $r_3$\}. $s_1$ and $s_2$ are two 
similar statements that differ by only a few words; one of them makes sense (i.e., conforms to common sense) while the other does not. They are used in our Subtask A: the \textbf{Validation} subtask, which requires a model to identify which one makes sense. For the statement that does not make sense, we have three candidate reasons, i.e., three options $o_1$, $o_2$, and $o_3$; one of them explains why the statement does not make sense. So, in our Subtask B, the \textbf{Explanation (Multi-Choice)} subtask, a model is required to find the correct reason from the three options. For the same nonsensical statement, in Subtask C, the \textbf{Explanation (Generation)} subtask, a participating system needs to generate the reason why it does not make sense. Three references, $r_1$, $r_2$, and $r_3$, are used for evaluating Subtask C. Below we give an example for each subtask, in which we introduce some notations we will use in the paper.

%\textbf{Task: \textit{Com}monsense \textit{V}alidation and \textit{E}xplanation (\textit{ComVE})}

%Here is an example of three subtasks: 

\begin{itemize}
    \item \textbf{Subtask A: Validation}
\end{itemize}

\quad Task: \textit{Select the statement of the two that does not make sense.}

\quad $s_1$: {\it John put a turkey into a fridge.}

\quad $s_2$: {\it John put an elephant into the fridge.}

\vspace{2mm}

\begin{tabular}{p{15cm}}
 In this example, $s_1$ is a sensical statement, also denoted as $s_c$, while $s_2$ is the nonsensical statement, which is also denoted as $s_n$. \\
\end{tabular}

\begin{itemize}
    \item \textbf{Subtask B: Explanation (Multi-Choice)}
\end{itemize}

\quad Task: Select the best reason that explains why the given statement does not make sense. 

\quad Nonsensical statement ($s_n$): {\it John put an elephant into the fridge.}

\quad $o_1$: {\it An elephant is much bigger than a fridge.} 

\quad $o_2$: {\it Elephants are usually white while fridges are usually white.}

\quad $o_3$: {\it An elephant cannot eat a fridge.}

\vspace{2mm}

\begin{tabular}{p{15cm}}
 In this example, the option $o_1$ is the correct reason, which is also denoted also as $o_c$, while $o_2$ and $o_3$ are not the reason, which are also denoted as $o_{n1}$ and $o_{n2}$. \\
\end{tabular}

\begin{itemize}
    \item \textbf{Subtask C: Explanation (Generation)}
\end{itemize}

\quad Task: Generate the reason why this statement does not make sense.

\quad Nonsensical statement ($s_n$): {\it John put an elephant into the fridge.}

\quad Reference reasons (used for calculating the BLEU score): 

\quad $r_1$: {\it An elephant is much bigger than a fridge.} 

\quad $r_2$: {\it A fridge is much smaller than an elephant.} 

\quad $r_3$: {\it Most of the fridges aren't large enough to contain an elephant.}

\vspace{2mm}

%\begin{tabular}{p{15cm}}
% Note that, the correct reason $o_c$ in Subtask B also repeat as one reference reason of Subtask C. \\
%\end{tabular}

\subsection{Evaluation Metrics}

The Subtasks A and B are evaluated using accuracy. Subtask C is evaluated with the BLEU score~\cite{papineni-etal-2002-bleu}. In addition, for Subtask C, we further perform human evaluation. We randomly select 100 instances from the test set and evaluate system outputs on Amazon Mechanical Turk. We ask three different crowd-sourcing workers to score each generated reason with a scale ranging from 0 to 3, inclusively, according the rubrics listed in Table~\ref{tab:rubrics}.

%Given a sentence that does not make sense together with a reason.
%a worker scores the appropriateness of the given reason on the scale [0,1,2,3], according to the rubrics below.

\begin{table}[t]
\begin{tabular}{| c | m{14cm} | }
\hline
{\bf Score} & {\bf Description} \\
\hline
 0 & The reason is not grammatically correct, or not comprehensible at all, or not related to the statement at all. \\ 
 \hline
 1 & The reason is just the negation of the statement or a simple paraphrase. Obviously, a better explanation can be made. \\  
 \hline
 2 & The reason is relevant and appropriate, though it may contain a few grammatical errors or unnecessary parts. Or like case 1, but it's hard to write a proper reason. \\
 \hline
 3 & The reason is appropriate and is a solid explanation of why the statement does not make sense. \\
 \hline
\end{tabular}
\caption{Rubrics used in human evaluation in Subtask C.}
\label{tab:rubrics}
\end{table}

Then we calculate the average score of the three scores as our final human evaluation score. Formally, the human evaluation score of system $k$ is
\begin{equation}
  score_k = \frac{\sum_{i=1}^{100} \sum_{j=1}^{3} score_{ijk}}{100 * 3},
\end{equation}
where $score_{ijk}$ means the score from the $j_{th}$ annotator for system $k$ on the $i_{th}$ instance.

\section{Data Construction}

Our data construction is mainly performed on Amazon Mechanical Turk, which consists of two steps:
\begin{itemize}
    \item Step 1: In this step, we construct datasets for Subtask A and Subtask B. Specifically, we ask a crowd-sourcing worker to write a sensical statement $s_c$ and a nonsensical statement $s_n$. For the nonsensical statement $s_n$, the worker further writes three sentences, $o_1$, $o_2$, $o_3$; one of them, denoted as $o_c$, explains why the nonsensical statement does not make sense; two of them, denoted as $o_{n1}$ and $o_{n2}$, serve as the confusing choices. (Refer to Section~\ref{sec:step1} for details.)
    \item Step 2: We then make three reference reasons, $r_1$, $r_2$, $r_3$ for Subtask C. We use $o_c$ as one of three references, and collect two more references in this step. We ask two different crowd-sourcing workers to write each of them. Note that instead of letting the same worker in step 1 to write these two references, we asked two more workers. The reason is to encourage diversity of the reference. (Refer to Section~\ref{sec:step2} for details.)
\end{itemize}
\indent Finally, each instance of the dataset have 8 sentences: \{$s_1$, $s_2$, $o_1$, $o_2$, $o_3$, $r_1$, $r_2$, $r_3$\}. Note that one sentence in $o_1$, $o_2$, $o_3$ is repeated in $r_1$, $r_2$, $r_3$, but for convenience of description, we denote it differently. 

\subsection{Step 1: Collecting Data for Subtask A and B}
\label{sec:step1}
%To be more convenient in the following section, we have marked abbreviations. Two statements $s_1$, $s_2$ can be classified into one sensical statement $s_c$, and one nonsensical statement $s_n$; Three option reasons  $o_1$, $o_2$, $o_3$ can be classified into one correct reason $o_c$, two confusing reasons $o_{n1}$ and $o_{n2}$; Three referential reasons $r_1$, $r_2$, $r_3$ can be classified into correct reason $o_c$, two additional referential reasons $r_{c1}$ and $r_{c2}$.

%The data is mainly collected with Amazon Mechanical Turk. For each instance, which includes one sensical statement $s_c$, one nonsensical statement $s_n$, one correct reason $o_c$, two confusing reasons$o_{n1}$, $o_{n2}$, and three referential reasons $o_c$, $r_{c1}$, $r_{c2}$, we separate the annotation process into two steps:
%\begin{itemize}
%    \item Step 1: We ask one crowd-sourcing worker to write one sensical statement $s_c$, one nonsensical statement $s_n$, one correct reason $o_c$, and two confusing reasons$o_{n1}$, $o_{n2}$. Step 1 is for Subtask A and Subtask B.
%    \item Step 2: We ask two other crowd-sourcing workers to write one correct reason $r_{c1}$ or $r_{c2}$ for each. Step 2 is for Subtask C.
%\end{itemize}

\paragraph{Annotation Guidelines.} When writing instances, workers were asked to follow several principles: 
(1) Try to avoid complex knowledge and focus on daily common sense. Make the questions as understandable as possible, so that a literate person is able to give the right answers.
(2) The confusing reason options, $o_{n1}$ and $o_{n2}$, should better contain more content words or information such as entities and activities in the nonsensical statements $s_n$. For example, the confusing reasons of  ``\emph{John put an elephant into the fridge}" should better contain both ``\emph{elephant}" and ``\emph{fridge}".
(3) The confusing reasons, $o_{n1}$ and $o_{n2}$, should be related to the statements $s_n$ and the correct reason $o_c$ and not deviate from the context; otherwise it may be easily captured by pretrained models like BERT \cite{COMMONSENSEQA}.
(4) The three option reasons, $o_1, o_2$, and $o_3$ should only be related to the incorrect statements $s_n$ rather than the correct statements $s_c$, because we want further studies to be able to estimate nonsensical statements $s_n$ without the correct statement $s_c$. 
(5) The confusing reasons, $o_{n1}$ and $o_{n2}$, should make sense themselves. Otherwise, the models may simply ignore the incorrect options $o_{n1}$, $o_{n2}$ without considering the casual semantics. This concern is raised from and motivated by the fact that models can achieve high performance in the ROC Story Cloze Task, when only looking at the alternative endings and ignoring the story content \cite{Schwartz2017}.
(6) We ask the annotators to make the nonsensical statement $s_n$ contain about the same number of words as the sensical statement $s_c$, and the correct reason $o_c$ have similar length with other two options. We drop the instances which do not meet such requirements.

%However, those principles are all soft restrictions; some instances are still good ones even without obeying those principles.  

\paragraph{Use of Inspirational Materials.} It is not easy for all crowd-sourcing workers to write instances from scratch. To address this issue, we also provide them with external reading materials to stimulate inspiration, such as the sentences of the Open Mind Common Sense (OMCS) project \cite{OMCS}. For example, \emph{``he was sent to a (restaurant)/(hospital) for treatment after a car crash"} can be inspired by the two sentences \emph{``restaurants provide food"} and \emph{``hospitals provide medical care"}.
%In addition, we include a small number of existing commonsense reasoning questions such as WSC \cite{WSC2012,WSC2015}, COPA \cite{COPA}, and SQUABU \cite{SQUABU}. 

%However, those inspirational sentences should not be used directly. 

% \begin{table}[t]
%   \begin{center}
%   \begin{tabular}{l|rrr}
%   \hline \bf type & \bf Training Data& \bf Dev Data & \bf Test Data \\\hline
%   Instances & 10,000 & 997 & 1,000\\
% %   Sensical Statement (Subtask A)& 10,000 & 997 & 1,000\\
% %   Nonsensical Statement (Subtask A\&B\&C)& 10,000 & 997 & 1,000\\
% %   Correct Reasons (Subtask B) & 10,000 & 997 & 1,000\\
% %   Confusing Reasons (Subtask B) & 20,000 & 1,994 & 2,000\\
% %   Referential Reasons (Subtask C) & 30,000 & 2,991 & 3,000\\
%   \hline
%   \end{tabular}
%   \end{center}
%   \vs.pace{-2mm}
%   \caption{\label{data_detail} Data amount }
% \end{table}

\paragraph{Quality Control.}
To ensure the quality of the data, we manually check the instances and drop or request a rewriting of the low-quality ones. If one worker writes too many low-quality instances, we will remove her or him from our annotator pool. With such process, we finally accept around 30\% submitted instances.

\subsection{Step 2: Collecting Data for Subtask C}
\label{sec:step2}
\paragraph{Annotation Guidelines.} To collect data for Subtask C, each worker is given a nonsensical statement $s_n$ and a sensical statement $s_c$ and asked to write a reason to explain why the nonsensical statement $s_n$ does not make sense. They shall follow the following rules:
(1) Do not explain why the sensical statement $s_c$ makes sense.
(2) Avoid mentioning the sensical statement $s_c$.
(3) Write the reason, rather than simply add the word \textit{``not"} or \textit{``can't"} to the nonsensical statement $s_n$ to form an explanation.
(4) Write the reason, don't use patterns like \textit{``XXX is not for YYY"} to create an explanation.
(5) Do not try to justify why the nonsensical statement $s_n$ makes sense.
(6) Write only one sentence, do not be overly formal.
(7) Refrain from using \textit{``because"} at the beginning of a sentence.
(8) Do not try to correct the statement $s_n$, but just give the reason.

\noindent{\bf Quality Control.}
As the same as in Step 1, after the annotators write the reasons in Step 2, the first two authors of the paper perform the check process again. We reject low-quality reasons (that violate the rules significantly) and low-quality annotators (who write many low-quality reasons with the number above a threshold).

\subsection{Data Summary and Analysis}
%As shown in Table ~\ref{data_detail}, f
For SemEval-2020, we created 11,997 instances (i.e., 11,997 8-sentence tuples). We further split the instances into three subsets with 10,000 (the training set), 997 (the development set), and 1,000 (the test set) instances, respectively. We randomly assign the label of the correct options in subtask A and B to avoid unbalanced correct labels.
%To ensure that development and test set have better quality, we have been more strict on the two sets. 
We conduct three more data analysis experiments to evaluate data quality, 
including sentence length, common words and repetition.

\begin{table}[t]
\small
  \begin{tabular}{l|ccc}
    \hline \hline
    \textbf{Type of Sentences }& \textbf{Training Set}& \textbf{Dev Set} & \textbf{Test Set}\\
    \hline
    Sensical Statements & 7.67 & 7.12 & 7.25 \\
    Nonsensical Statements & 7.69 & 7.16 & 7.36\\
    Correct Reasons & 8.13 & 7.96 & 8.09\\
    Confusing Reasons & 7.80 & 7.14 & 7.29\\
    Referential Reasons &8.08& 7.92& 8.06\\
    \hline
    \end{tabular}
    \caption{Average length of different types of sentences of Training/Dev/Test set}
    \label{average_length}
\end{table}

\noindent{\bf Average Length.}
In Table~\ref{average_length}, we present the average length of each type of sentence in the training/dev/test set. 
The sentences in the development and test set have shorter lengths than those in the training set. This is because we check the development and test more carefully and more strictly, thus removing longer and more  incomprehensible instances, which lowers the average lengths of the dev/test set. The sensical statements and nonsensical statements almost have the same average lengths in the three sets (the differences are equal or smaller than 1\%), which is balanced. However, there is an obvious gap between the correct reasons and confusing reasons in terms of the average lengths (roughly 4\% in the training set and 10\% in the dev/test set).

%, which will cause some improper hints for models.

\begin{table*}[t]
  \centering
  \small
  \setlength{\tabcolsep}{1mm}
  \begin{tabular}{c|ccccc}
  \hline \hline
  \textbf{Types of Sentences} 
  & \multicolumn{5}{c}{\textbf{ Word:Word Frequency(\textperthousand)}} \\
  \hline
  \makecell[c]{Sensical Statements} & can:1.54&his:1.45&with:1.285&my:1.247&people:0.912  \\
  \makecell[c]{Nonsensical Statements} & can:1.604&his:1.411&with:1.299&my:1.254&people:0.873\\
  \makecell[c]{Correct Reasons} & not:4.545&cannot:1.731&people:1.579&can:1.389&no:1.337\\
  \makecell[c]{Confusing Reasons} & can:2.095&people:1.671&not:1.49&have:1.152&than:0.81\\
  \makecell[c]{Referential Reasons} &not:5.711&cannot:1.65&can:1.09&people:1.088&have:0.978\\
\hline
\multicolumn{6}{c}{(a) Training set} \\ 
\hline \hline
  \textbf{Types of Sentences} 
  & \multicolumn{5}{c}{\textbf{ Word:Word Frequency(\textperthousand)}} \\
  \hline
  \makecell[c]{Sensical Statements} &  can:1.343 & my:1.322 & his:1.291 & put:1.259 & with:0.978 \\
  \makecell[c]{Nonsensical Statements} &  can:1.458 & my:1.335 & put:1.304 & his:1.15 & with:0.975 \\
  \makecell[c]{Correct Reasons} &  not:2.325 & can:1.709 & no:1.614 & people:1.362 & than:1.31 \\
  \makecell[c]{Confusing Reasons} &  can:2.332 & people:1.619 & not:1.609 & have:1.094 & than:1.079 \\
  \makecell[c]{Referential Reasons} &  not:4.264 & cannot:1.212 & can:1.203 & it:1.179 & have:1.093 \\
\hline
\multicolumn{6}{c}{(b) Dev+Test set} \\ 
  \end{tabular}
  \vspace{-2mm}
  \caption{Top-5 common words and their frequencies in different types of sentences in the training and dev+test set. 1.000\textperthousand \ means this word appear once in every 1000 words. 
  }
  \label{common_words}
\end{table*}

\noindent{\bf Common Word Analysis.}
The most common words are important for showing the differences between sentences. We only present those words which have obvious different frequencies between sensical statements and nonsensical statements or between correct/referential reasons and confusing reasons. So, we skip most uninformative words, including `a', `an', `the', `to', `in', `on', `of', `for', `and',  `is', `are' and `be'.  After removing those words, we can list the top-5 common words in each type of sentence in the training/dev+test sets. 
For sensical statements $s_c$ and nonsensical statements $s_n$, there are no significant differences between the training, dev, and test set. 
However, there is an obvious gap in the correct reasons $o_c$ and confusing reasons $o_n$ in negative words such as ``not", ``no", and ``cannot". In the training data, negative words are about 3 times more common in the correct option $o_c$ than in the confusing options $o_n$. In the dev+test data, the gap is about 40\%, which indicates that the dev+test data has a higher quality than the training data. However, as discussed in \cite{Probing}, \textit{spurious statistical cues} can affect BERT's results. We conjure that the negative words are also spurious effective clues, which make the Subtask B potentially easier.

\noindent{\bf Repetition.}
The dev+test set have 12 instances (0.6\%) that repeat the same nonsensical statements in the training data and 36 instances (1.8\%) that repeat the same correct reasons with the training data.

\subsection{Cautions of using the data} The following advice is given to all task participants and future users:
(1) Feel free to use whatever additional data they deem appropriate for the tasks to train their model. (2) Do not use the input of Subtask B/C to help Subtask A and do not use the option $o$ of Subtask B to help Subtask C. Otherwise the task will be artificially easy. This is because of two reasons: a) The nonsensical statements $s_n$ of Subtask B and Subtask C is exactly the nonsensical statements $s_c$ of Subtask A and, participants can use the input of the Subtask B/C to directly obtain the answer of Subtask A and the option answers $o$ of Subtask B will also reduce the difficulty of Subtask A; b) the correct reason $o_c$ of Subtask B is also one of the reference reason $o_c$ in Subtask C.

% {\bf Evaluation Scheduling} The test period of each subtask is held at different weeks because the statement of Subtask B/C is the answer of Subtask A, one reference reason of Subtask C is the answer of Subtask B. So, during the evaluation period, we first held Subtask A, then Subtask C and last Subtask B, one week for each subtask. We schedule the test period like this is because 1. the input of Subtask B and Subtask C is exactly the output of Subtask A, participants can use the input of the Subtask B/C to directly get the answer of Subtask A; 2. the correct reason of Subtask B is also the one referential reason of Subtask B, participants can use the correct answer of Subtask B to answer Subtask C. 

\section{Systems and Results}

In this section, we show the evaluation results of all the submitted systems for the three subtasks. Since most systems share similar model architecture for subtasks A and B, we discuss the two subtasks together. 

\subsection{Subtask A and Subtask B}

\begin{figure}[t]
	\centering
	\includegraphics[width=\textwidth]{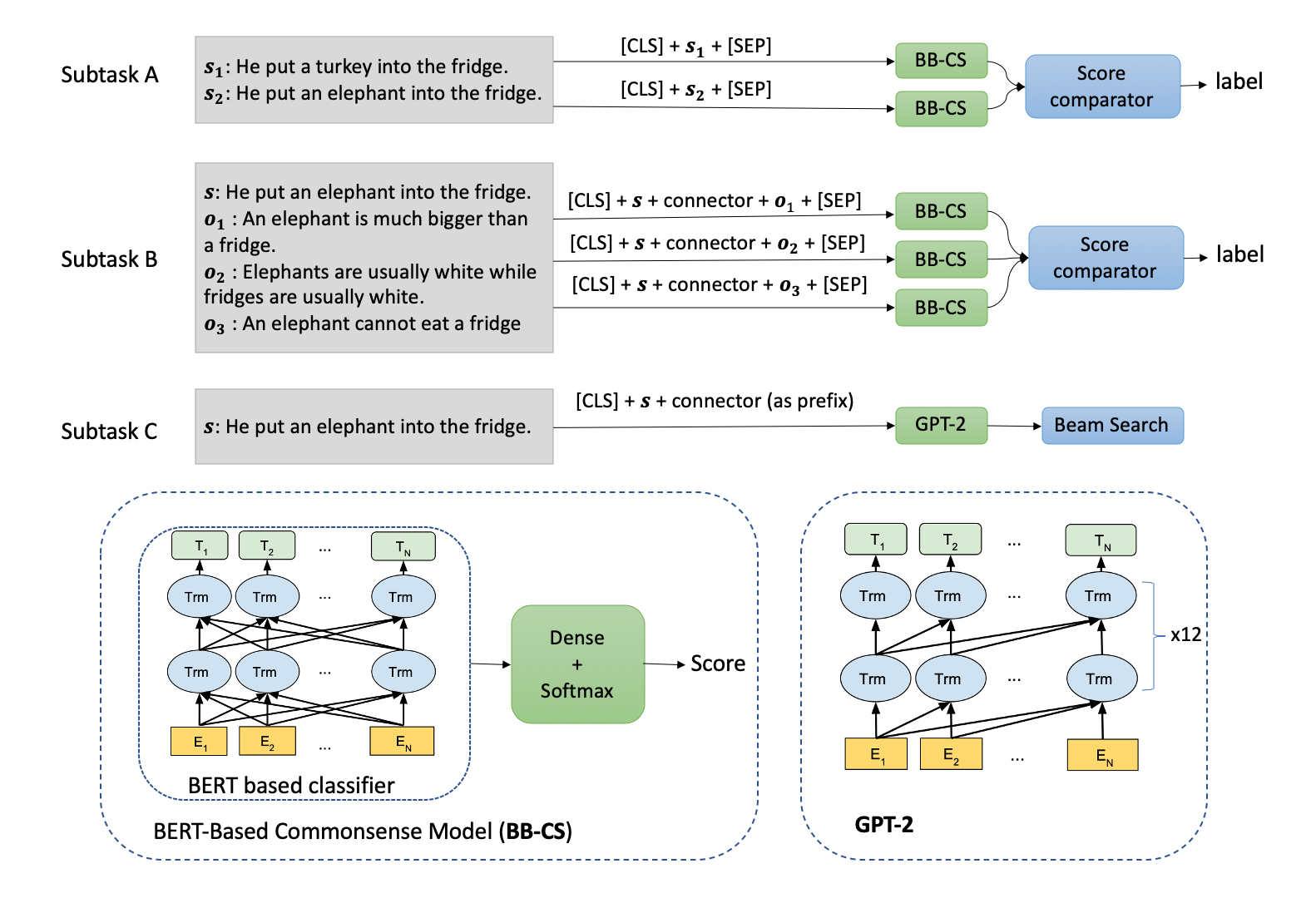}
	\caption{The most commonly used model architectures used in the three subtasks. This figure is mostly based on Team Solomon's system. For Subtask B and C, the connector can be simply ``No, ", to help in constraining the model to learn a choice that explains the unreasonability of the statement. For Subtask A and B, the pretrained models are finetuned on the task-specific data with MLM-objective, and then trained as a binary classification task to score each input. For Subtask C, the cross-entropy loss of next-token-prediction is used to train the model, and beam search is used at inference.}
	\label{fig:models}
\end{figure}

\begin{table}[t]
  \small
  \begin{tabular}{llc|llc|llc}
  \hline\hline
  {\textbf{Team}} & {\textbf{Acc.}} & {\textbf{Rank}} & {\textbf{Team}} & {\textbf{Acc.}} & {\textbf{Rank}} & {\textbf{Team}} & {\textbf{Acc.}} & {\textbf{Rank}} \\
  \hline
  Human & {\it 99.1} & - & \multicolumn{5}{c}{}  & \\
  \hline
  {CN-HIT-IT.NLP }         & {97.0} &1    & {panaali*}            & {92.5}  &14            & {Lijunyi}            & {83.0}   &27           \\
  {ECNU-SenseMaker}          & {96.7} & 2             & {ZhengxianFan*}       & {92.4}  &15            & {ehsantaher*}         & {82.5}  &28            \\
  {IIE-NLP-NUT}                & {96.4} & 3             & {LMVE}              & {90.4} &16             & {TakeLab*}            & {81.2}     &29         \\
  {nlpx*}               & {96.4}  &3            & {Warren*}             & {90.4}  &16            & {Vicki*}              & {79.8}  &30            \\
  {Solomon}            & {96.0}  &5            & {TMLab*}              & {89.2}   &18           & {TR}                 & {79.7}   &31           \\
  {Qiaoning}           & {95.9}  &6            & {UAICS}              & {89.1}  &19            & {KDE SenseForce}     & {79.6}   &32           \\
  {BUT-FIT}            & {95.8} &7             & {JUST}               & {89.1}  &19            & {Hitachi*}            & {78.4}   &33           \\
  {olenet*}             & {95.5} &8             & {eggy*}               & {89.0} &21              & {CUHK}               & {72.4}  &34            \\
  {KaLM}           & {95.3}  &9            & {UI}   & {88.2}     &22           & {paramitamirza*}      & {69.2} &35           \\
  {CS-NET}         & {94.8}  &10            & {Armins*}             & {87.1}   &23           & {UoR}                & {67.6}   &36           \\
  {fkerem*}             & {94.4}  &11            & {DEEPYANG}        & {85.1}  &24            & {chenggguang*}        & {62.3}  &37            \\
  {JUSTers}            & {92.9}  &12            & {WUY*}                & {84.2}    &25          & {praveenjoshi007*}    & {55.9} &38             \\
  {CS-NLP}        & {92.7} &13             & {YNU-oxz}           & {83.6}     &26         & {dania*}              & {21.6}  &39            \\
     \hline
\end{tabular}
\caption{Subtask A results of all the submitted systems. Those marked with * did not submit system description paper. Human performance are based on the trial data instead of the test data used by the participating systems.}
\label{tab:Subtask A_results}
\end{table}

\begin{table}[t]
  \small
  \begin{tabular}{llc|llc|llc}
    \hline\hline
  {\textbf{Team}} & {\textbf{Acc.}} & {\textbf{Rank}} & {\textbf{Team}} & {\textbf{Acc.}} & {\textbf{Rank}} & {\textbf{Team}} & {\textbf{Acc.}} & {\textbf{Rank}}\\
  \hline
  Human & {\it 97.8} & - & \multicolumn{5}{c}{}  & \\
  \hline
  {ECNU-SenseMaker}          & {95.0} &1             & {JBNU}               & {91.4}   &10           & {Masked Reasoner}        & {73.5}       &19       \\
  {CN-HIT-IT.NLP}         & {94.8} &2             & {Qiaoning}           & {90.8}   &11           & {KDE SenseForce}     & {72.8} &20             \\
  {IIE-NLP-NUT}                & {94.3}   &3           & {CS-NET}         & {89.0}  &12            & {SSN-NLP}           & {68.3}      &21        \\
  {Solomon}            & {94.0}   &4           & {WUY*}                & {85.3}    &13          & {TakeLab*}            & {66.8}       &22       \\
  {LMVE}              & {93.8} &5             & {SWAGex}               & {84.6}  &14            & {UoR}                & {65.9}      &23        \\
  {CS-NLP}        & {93.7}  &6            & {TMLab*}              & {82.0}    &15          & {dania*}              & {55.5}    &24          \\
  {KaLM}           & {93.2}  &7            & {UI}   & {80.5}      &16        & {CUHK}               & {51.2}   &25           \\
  {BUT-FIT}            & {93.1}  &8            & {ehsantaher*}         & {79.3}   &17           & {bhu*}                & {36.4}  &26            \\
  {JUSTers}            & {92.3}  &9            & {uzh*}                & {75.8} &18             & {praveenjoshi007*}    & {32.6}     &27         \\
  \hline
  \end{tabular}
  \caption{Subtask B results of all the submitted systems. Those marked with * did not submit system description paper. Human performance are based on the trial data instead of the test data used by the participating systems.}
  \label{tab:Subtask B_results}
  \end{table}

The formal evaluation results of Subtask A and B are shown in Table~\ref{tab:Subtask A_results} and \ref{tab:Subtask B_results}.
There are in total 39 valid submissions for Subtask A and 27 valid submissions for Subtask B.  Most top-performing submissions adopted the pretrained language models such as BERT~\cite{BERT}, RoBERTa~\cite{RoBERTa}, XLNET~\cite{yang2019xlnet} and ALBERT~\cite{lan2019albert} as the encoder of the model, and then finetune on the training set of the task. See Figure \ref{fig:models} for the most commonly-used model architectures for Subtask A and B. Also, the top-performing systems take advantage of external knowledge graphs such as ConceptNet~\cite{ConceptNet5.5}, or unstructured text containing commonsense knowledge. Below we introduce in detail several top-performing systems and their main features. 

\begin{itemize}
  \item {\bf CN-HIT-IT.NLP}~\cite{CN-HIT-IT.NLP} ranks top in Subtask A. They use a variant of K-BERT ~\cite{liu2019k} as the encoder to enhance language representations through knowledge graphs. K-BERT is a Transformer-based model, which enhances the language representations of the text by injecting relevant triples from a knowledge graph to form a knowledge-rich sentence tree, and then uses a mask-Transformer to make the triples visible only to the corresponding entity. They use ConceptNet as the commonsense repository to extract the triples for the statements.
  \item {\bf ECNU-SenseMaker}~\cite{ECNU-SenseMaker} ranks top in Subtask B. They use Knowledge-enhanced Graph Attention Network to leverage heterogeneous knowledge from both the structured knowledge base (i.e. ConceptNet) and the unstructured text to better improve the commonsense understanding. Like {\bf CN-HIT-IT.NLP}, their model is also based on K-BERT. In addition, they use unstructured text from ConceptNet and Subtask C to pretrain the language model.
  \item {\bf IIE-NLP-NUT}~\cite{IIE-NLP-NUT} uses RoBERTa as the encoder, and conduct a second pretraining on the original RoBERTa model with the textual corpus from Open Mind Common Sense~\cite{singh2002open}. They also explore several prompt templates to constructs as the inputs to the model.
  \item {\bf Solomon}~\cite{Solomon}, {\bf KaLM}~\cite{KaLM}, {\bf CS-NET}~\cite{CS-NET}, {\bf JUSTers}~\cite{JUSTers}, {\bf CS-NLP}~\cite{CS-NLP}, {\bf UI}~\cite{UI}, {\bf TR}~\cite{TR} {\bf UoR}~\cite{UoR}, {\bf Masked Reasoner}~\cite{MaskedReasoner} have similar model architecture, with RoBERTa as the encoder.  In addition, {\bf UoR} finetunes the pretrained language model on NLI and STS dataset, and {\bf UI} finetunes on MNLI data. {\bf TR} combines RoBERTa features with additional features from text-to-image generation using Gradient Boosted Decision Tree, and give better results in post-evaluation.
  \item {\bf Qiaoning}~\cite{QiaoNing} and {\bf JUST}~\cite{TeamJUST} use several ensembles of BERT, ALBERT, XLNet and RoBERTa.
  \item {\bf BUT-FIT}~\cite{BUT-FIT}, {\bf LMVE}~\cite{LMVE}, {\bf Lijunyi}~\cite{Lijunyi} use ALBERT as the encoder. {\bf BUT-FIT} uses back-translation from Czech for data augmentation, and {\bf LMVE} uses hint sentences, back-translation from French and intra-subtask transfer learning between Subtasks A and B to enhance their system.
  \item {\bf UAICS}~\cite{UAICS}, {\bf DEEPYANG}~\cite{DEEPYANG}, {\bf YNU-oxz}~\cite{YNU-oxz}, {\bf KDE-SenseForce}~\cite{KDE}, {\bf CUHK}~\cite{CUHK} {\bf JBNU}~\cite{JBNU}, {\bf SWAGex}~\cite{SWAGex} are BERT-based. {\bf JBNU} puts an BiLSTM on top of BERT, and {\bf SWAGex} finetunes BERT with SWAG data.  {\bf CUHK} uses a Multitask Learning framework MTDNN~\cite{liu2019multi}, adopting the ``Explain, Reason and Predict" system.
\end{itemize}

It can be seen from the results that pretrained language models such as RoBERTa can achieve rather high performance, e.g., the team Solomon achieves 96.0\% and 94.0\% on Subtask A and Subtask B, respectively, without using further resources. This shows that large-scale pretrained language models do contain commonsense knowledge to deal with the Subtask A and the Subtask B in this challenge. Additionally finetuning the pretrained language models on commonsense-related text such as OMCS, which we use as inspirational materials, can push the results even higher, close to human performance. The best-performing teams on Subtask A and Subtask B both adopt K-BERT, which incorporates the external knowledge base (i.e. ConceptNet) to complement the pretrained language models with knowledge triples. This shows that knowledge-graph-enhanced approaches, such as K-BERT can effectively incorporate external knowledge. However, the high number may also indicate data leaking to some extent, since in the data creation stage, both ConceptNet and OMCS are used as references for the annotator to write the data instances.

\subsection{Subtask C}

\begin{table}[t]
  \small
  \begin{tabular}{l|cc|cc||l|cc|cc}
  \hline\hline
  {\textbf{Team}} & {\textbf{BLEU}} & {\textbf{Rank}} & {\textbf{Human}} & {\textbf{Rank}} &  {\textbf{Team}} & {\textbf{BLEU}} & {\textbf{Rank}} & {\textbf{Human}} & {\textbf{Rank}}  \\
  \hline
  Human & - & - & 2.58 & - &  \multicolumn{3}{c}{}  & \\
  \hline
  {BUT-FIT}  & {22.4} & 1  & {1.84}  & 4                         & {CN-HIT-IT.NLP+}       & {9.7}   &10        & {1.74} & 10                            \\
  {Solomon}       & {19.3} & 2         & {1.84}  & 4                          & {SWAGex}             & {7.1} &11          & {1.75}  & 8                          \\
  {KaLM}      & {18.5}   & 3     & {2.08}  & 2                           & {UI} & {5.5}   &12        & {0.73}  & 14                           \\
  {panaali*}       & {17.2}  & 4        & {1.22}    & 11                        & {TMLab*}        & {5.4}      &13         & {1.05}    & 12                        \\
  {JUSTers}       & {16.1}  &5        & {1.94}  & 3                          & {CUHK}        & {4.3}   &14            & {0.58}     &16                       \\
  {cdjhz*}         & {16.0}  &6        & {1.75} &8                           & {SSN-NLP}      & {2.2}     &15         & {0.59}         &15                   \\
  {JBNU}          & {15.9} &7         & {1.80}   & 6                         & {UoR+}             & {0.9}  &16          & {0.53}           &17                 \\
  {ANA}           & {15.7}    &8      & {2.10}   & 1                         & {Masked Reasoner+}       & {0.6}   &17        & {0.81}      &13                      \\
  {LMVE+}         & {12.9}    &9      & {1.78}   & 7                         & {}                 &                                      &              \\
  \hline                                          
  \end{tabular}
  \caption{Subtask C results of all the submitted systems. Those marked with * did not submit a system description paper, and those marked with + means they do not include Subtask C in their system description paper.}
  \label{tab:Subtask C_results}
\end{table}

The results for Subtask C are shown in Table~\ref{tab:Subtask C_results}. There are in total 17 valid submissions for Subtask C. There are generally two approaches: (1) sequence-to-sequence approach, where the source side is the non-sensical statement, and the reason is the target sequence. (2) language model generation approach, which uses large-scale pretrained auto-regressive language models such as GPT-2~\cite{radford2019language} for reason generation, where the non-sensical sentence acts as prompt. An example of the language model generation approach is shown in Figure~\ref{fig:models}, which is most commonly used and achieves relatively good results. Below we describe in detail the systems and their main features.

\begin{itemize}
  \item {\bf BUT-FIT}~\cite{BUT-FIT} experiments with both the sequence-to-sequence approach and the language generation approach. For the sequence-to-sequence approach, they use BART~\cite{lewis2019bart} with beam-search decoding to achieves the highest BLEU among all the teams. For the language generation approach, the nonsensical statement is used as a prompt. At the training stage, the statement and the explanation are concatenated together, and a GPT-2 is trained on these sequences with a next token prediction objective. At the test time, based on the statement, the model generates the reason tokens until the end-of-sentence token is generated.
  \item {\bf KaLM}~\cite{KaLM} uses the sequence-to-sequence architecture BART. To enhance the source side statement, they extract keywords from the statement and search for evidence from Wiktionary.\footnote{Wiktionary version: enwiktionary-20200220} After that, they concatenate the evidence along with the original statement as the source sentence for the generation. This approach proves effective and makes their system second-best for human evaluations.
 
 \item {\bf ANA}~\cite{ANA} has the highest human evaluation score with a multitask learning framework. Specifically, they use a decoder-only transformer based on GPT-2 as the backbone model, and train the model with two self-attention heads: one for language models and another for classification. They then use data from both task B and task C to calculate language model loss and classification loss. Furthermore, they use OMCS at the pretraining stage and use CoS-E~\cite{rajani2019explain} and OpenBook~\cite{OpenBookQA} at the task-specific training stage.

  \item {\bf Solomon}~\cite{Solomon}, {\bf JUSTers}~\cite{JUSTers}, {\bf SWAGex}~\cite{SWAGex}, {\bf UI}~\cite{UI} and {\bf CUHK}~\cite{CUHK} use GPT or GPT-2 finetuned on the task training data. {\bf JBNU}~\cite{JBNU} uses UniLM, which incorporates three LM tasks: unidirectional LM, bidirectional LM and sequence-to-sequence prediction LM, and only use one of the reference correct reasons. {\bf UI} does not use the training data and treats the generation as a Cloze task. {\bf SSN-NLP}~\cite{SSN-NLP} uses the seq2seq NMT framework without a pretrained LM.

\end{itemize}

Large-scale pretrained language models such as BART and GPT-2 dominates the submissions. The two systems with the highest human evaluations, namely {\bf ANA} and {\bf KaLM}, use additional resources such as Wiktionary, OMCS, and other commonsense datasets. This again shows that additional knowledge from structured databases can help with the generation of the reasons. From Table~\ref{tab:Subtask C_results} we can see that BLEU does not correlate well with Human Evaluation, especially for the top-performing systems. According to a further experiment of BUT-FIT, the naive baseline of ``copying source sentence as the reason" can give a BLEU of 17.23, which can rank No. 4 among all the submissions. This indicates that BLEU, which focuses on the surface token overlap, has difficulty in evaluating the generated text reliably. The top-performed system achieves the human evaluation score of 2.10, showing the power of pretrained language models, but considering the human performance of 2.58, we still have a long way to go to generate human acceptable reasons.

% \begin{table*}
%   \begin{floatrow}
%   \capbtabbox{
%    \begin{tabular}{l|cc}
%    \hline
%    Dataset & First Statement Invalid& Second Statement Invalid\\
%    \hline
%    Training Set & 4979 & 5021 \\
%    Dev Set & 518 & 479 \\
%    Test Set & 492 & 508 \\
%    \hline
%    \end{tabular}
%   }{
%    \caption{Label Distribution for Subtask A}
%    \label{tab:tb1}
%   }
%   \capbtabbox{
%    \begin{tabular}{l|rrr}
%    \hline
%    Dataset & OptionA correct & OptionB correct & OptionC correct\\
%    \hline
%    Training Set & 3195 & 3362 & 3443\\
%    Dev Set & 344 & 327 & 336 \\
%    Test Set & 320 & 355 & 325 \\
%    \hline
%    \end{tabular}
%   }{
%    \caption{Label Distribution for Subtask B}
%    \label{tab:tb2}
%   }
%   \end{floatrow}
%   \end{table*}
\section{Related Work}
Commonsense reasoning in natural language has been studied in different forms of tasks and has recently attracted extensive attention. In the Winograd Schema Challenge (WSC) \cite{WSC2012,WSC2015}, a model needs to solve hard co-reference resolution problems based on commonsense knowledge. For example, \emph{``The trophy would not fit in the brown suitcase because it was too big. What was too big (trophy or suitcase)?"}
%Given two sentences which must contain the same pronounce and only differ in a small part, the task asks to select the correct instance from two alternatives to replacing the pronounce. Each instance will fit the different sentences of the two. 
%However, WSC estimates common sense indirectly and does not consider an explanation of why one option is correct while the other is incorrect. 
The Choice of Plausible Alternatives (COPA) \cite{COPA} emphasizes on events and consequences. Each question in COPA aims to find the suitable cause or result of the premise from two given alternatives. All premises and alternatives are simple sentences. For example, the premise can be \emph{``The man broke his toe. What was the CAUSE of this?"} and the two candidate answers are \emph{``(1) He got a hole in his sock."} and
\emph{``(2) He dropped a hammer on his foot."}
Several subsequent datasets are inspired by COPA. The JHU Ordinal Common-sense Inference (JOCI) \cite{JOCI} aims to label the plausibility from 5 (very likely) to 1 (impossible) of human response after a particular situation. Situations with Adversarial Generations (SWAG) \cite{SWAG} request a system to choose the most likely-to-happen alternative after a specific situation. %Thanks to crowd-sourcing, the recent datasets are much larger than COPA. 
Those datasets emphasize the pre-situations and/or the after-situations of certain situations, but not on the reasons why they occur or are caused. Besides, our dataset is not limited to events or situations. It concerns a broader commonsense setting, which includes events, descriptions, assertion etc.

%有些地方用原论文的原话，可以吗？

Some datasets are inspired by reading comprehension. The Story Cloze Test and ROCStories Corpora \cite{RocStories2016,RocStories2018} aim to figure out the right ending from two candidate sentences after a four-sentence story. For a narrative text, MCScript \cite{MCScript} gives various types of questions and pairs of answer candidates for each question. Most questions require knowledge beyond the facts mentioned in the text. Compared to those reading comprehension tasks, our benchmark encourages people to use any external resources they want.

Some other datasets evolve from QA problems and care more about factual commonsense knowledge. SQUABU \cite{SQUABU} provides a small hand-constructed test of commonsense and scientific questions. CommonsenseQA \cite{COMMONSENSEQA} asks crowd workers to create questions from ConceptNet \cite{ConceptNet5.5}, which is a large graph of commonsense knowledge, where each question discriminates its answer candidates between three target concepts that all share the same relationship to a single source drawn from ConceptNet.
OpenBookQA \cite{OpenBookQA} provides questions and answer candidates, as well as thousands of diverse facts about elementary level science that are related to the questions. The AI2 Reasoning Challenge (ARC) \cite{ARC} gives thousands of questions with different knowledge types,
% e.g. Definition, Basic Facts \& Properties, 
as well as a relevant 14M-sentence corpus, mixed with science facts and other narrative sentences. MuTual provides a dataset for Multi-Turn dialogue reasoning in the commonsense area \cite{mutual}.
Those questions are not easy to answer without specializing certain domain knowledge, while our questions are based on daily common sense. 

Some datasets focus on non-sentential eventual plausibility \cite{wang2018modeling,porada2019can}, such as \textit{``gorilla-ride-camel"}. In contrast,  our dataset is based on \textit{statements} which includes events, descriptions, assertion etc, not merely events, such as \textit{``China's territory is larger than Japan's"}. And some datasets concentrate on limited attributes or actions of world knowledge, such as physics \cite{verbphysics}. Our dataset concerns general commonsense knowledge beyond just physical common sense, the sentence in our task \textit{``Tom's mom become (happy)/(upset) when Tom gets high grades in the exam"} is about social and emotional common sense. 
For our first task, those statements that conforms to commonsense can also be phrased as being plausible. Thus our first task is similar to plausibility tests, despite that plausibility has a broader scope while our focus is on commonsense only.

More importantly, compared with our work, the \textbf{above} tasks do not directly estimate general common sense or ask the logical reasons behind the correct answers and questions. In recent years, some large-scale commonsense inference knowledge resources have been developed, which may be helpful in commonsense reasoning tasks.
Atomic \cite{ATOMIC} presents a large-scale everyday commonsense knowledge graph, which has nine \emph{if-then} relations with variables, including causes, effects, and so on. 
Event2Mind \cite{Event2Mind} proposes a new corpus and task, aiming to find out the mentioned/unmentioned people's intents and reactions under various daily circumstances. These datasets are not directly useful for our benchmark since they focus only on a small domain. ConceptNet is a seminal knowledge graph that has been upgraded over time \cite{ConceptNet,ConceptNet3,ConceptNet5,ConceptNet5.5}. ConceptNet constructs triples using labeled edges as relations and various words and/or phrases as entities. It also has the sentences describing the corresponding triples. In contrast to these datasets, we investigate the evaluation of common sense, rather than building a resource. %Thus we consider using ConceptNet to help crowd-sourcing workers write instances.

Before organizing this shared-task, a pilot study \cite{wang-etal-2019-make} has been performed, showing that there is still a significant gap between human and machine performance when no training data is provided, despite that the models have already been pretrained with over 100 million natural language sentences. In our task here, we also provide training data with human annotations.

\section{Summary}
This paper summarizes SemEval-2020 Task 4: Commonsense Validation and Explanation. In this task, we construct a dataset that consists of 11,997 instances and 83,986 sentences. The task attracted around 40 participating teams, out of which 31 teams submit their system papers. 
The pretrained models are shown to be very effective in Subtask A and Subtask B, but there is still a large room to improve system performances in Subtask C. Contextualized embedding such as RoBERTa and BART play a central role in the success of the top-performing models, demonstrating that such methods contain commonsense information to a good extent.

%In this competition, we can see that models, especially pretrained models like RoBERTa \cite{RoBERTa} can perform well on Subtask A and Subtask B. 
We attribute the high performance on Subtask A and B to several main reasons: 
1) Subtask A is a relatively easy question by definition: a model needs only to detect a relatively less plausible content among the two candidate sentences. 
2) Pretrained models are obtained on billion-words large corpora such as Wikipedia data, which help obtain commonsense knowledge \cite{zhou2019evaluating}, which helps achieve considerably better performance. 
3) As described in the annotation process, we use the sentences from OMCS to inspire crowd-sourcing workers. The top-3 systems also use OMCS, which potentially help them to attain better performances. 
4) For Subtask B, as discussed in our data analysis section, the data has some flaws in the average length and common words, which reduces the difficulty. 
5) Some instances have obvious patterns. For example, there are tens of instances that contain ``put XXX into YYY", and ``XXX is bigger than YYY", making the problems simpler. 
6) Hundreds of crowd-sourcing workers write instances. It is likely for workers to think about the shared commonsense knowledge, such as ``XXX is bigger/shorter/quicker/slower than YYY". 

We consider future works in four directions: 
1) We observe that there is still a gap between machine performance and human performance in Subtask C, and the reason generation task still needs further investigation. 
2) The artifacts or spurious correlations in the datasets can be further removed, e.g., by making different candidate sentences in subtask B be the same, removing instances with shared commonsense knowledge, removing artifacts in common words, and filtering out common patterns. 
3) Subtask A can be turned into a more difficult form. Instead of comparing which statement makes more sense, we can form it into a classification task, validating if one statement makes sense or not. 
4) We notice that the BLEU score does not closely align with human evaluation for systems with high performances, and it is desirable to develop an auto-metric for comparing the semantic correlation between two reasons.

\section*{Acknowledgements}
This work is supported by the National Science Foundation of China (Grant No. 61976180), the Westlake University, and the Bright Dream Joint Institute for Intelligent Robotics. The research of Xiaodan Zhu is supported by NSERC. Yue Zhang is the corresponding author. 

% \begin{table*}
%   \centering
%   \begin{tabular}{c|ccc}
%   \hline

%   {\textbf{Dataset}} & \makecell[c]{Correct Reasons} &\makecell[c]{Confusing reasons} &  \makecell[c]{Balanced Number\\ Should be}\\ 
%   \hline
%   Train data &4837&5163& 3333:6667\\
%   Dev data  & 556 &444 &333:6667 \\
%   Dev data  & 552 &448 &333:6667 \\
%   \hline
%   \end{tabular}
%   \caption{\label{dataset-details}
%     }
% \end{table*}

% \begin{table*}
%   \centering
%   \begin{tabular}{c|cc}
%   \hline

%   {\textbf{Dataset}} & \makecell[c]{Sensical Statements} &\makecell[c]{Nonsensical Statements}\\ 
%   \hline
%   \makecell[c]{Direct Overlap\\ between \\Dev and Train} &{20-25\%}&{15-20\%}\\
%   \hline
%   \makecell[c]{Direct Overlap\\ between \\Test and Train} &{20-25\%}&{15-20\%}\\
%   \hline
%   \end{tabular}
%   \caption{\label{dataset-details}
%     }
% \end{table*}

\bibliographystyle{coling}
\bibliography{semeval2020}

\end{document}